\documentclass[letterpaper, 10 pt, conference]{ieeeconf}  

\IEEEoverridecommandlockouts                              

\usepackage{graphicx} 
\usepackage{amsmath} 
\usepackage{amssymb}  
\usepackage{xcolor}
\usepackage{cite}
\usepackage{booktabs}
\usepackage{dblfloatfix}

\usepackage{etoolbox}
\makeatletter
\patchcmd{\@makecaption}
  {\scshape}
  {}
  {}
  {}
\makeatother

\title{\LARGE \bf
DroneDAR: Long-Range Drone Distance Estimation Using Monocular Vision and Bounding-Box Features
}
\author{Knut Peterson$^{1*}$, Zaid Mayers$^{1*}$ and David Han$^{1}$
\thanks{$^{1}$iMaPLe Research Lab, Drexel University, Philadelphia, PA}
\thanks{\tt \{kp3275, zrm33, dkh42\}@drexel.edu}
\thanks{$^{*}$Equal Contribution}%
}

\begin{document}

\maketitle
\thispagestyle{empty}
\pagestyle{empty}

\begin{abstract}
Accurate distance estimation for small drones in long-range imagery is important for tracking and situational awareness, yet remains challenging due to extreme target scale variation, background clutter, and noisy visual cues. This paper studies monocular drone distance estimation using image crops together with bounding-box geometry, a practical setting in which a detector provides a candidate drone region and the model predicts range from appearance and box-derived features. We evaluate a Droneranger-style baseline, and introduce a new DroneDAR (\textbf{Drone} \textbf{D}etection \textbf{A}nd \textbf{R}anging) model that combines a convolutional backbone with explicit bounding-box cues through a lightweight gating mechanism. Experiments analyze how backbone capacity, crop resolution, and regression loss functions affect performance across distance regimes. We further examine common failure modes at long distances, including sensitivity to bounding-box noise and reduced texture detail in the crop. The results provide guidance for designing and training range estimators that remain robust under real-world long-range conditions and highlight directions for improving reliability when drones occupy only a few pixels.
\end{abstract}



\section{Introduction}

\begin{figure*}[tbp] 
\centering
\includegraphics[width=0.99\textwidth]{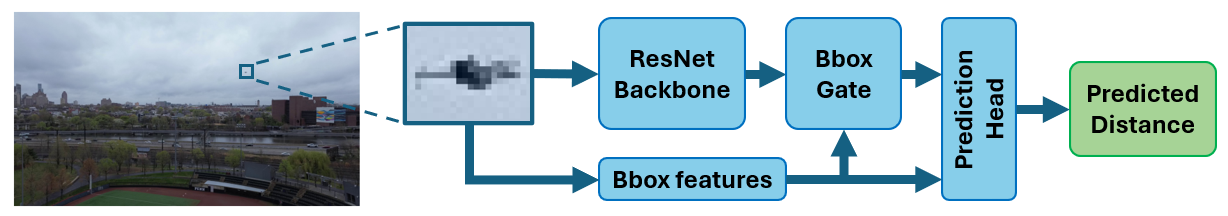}
\caption{Proposed model architecture for predicting the distance of a drone from the camera. We build on the DroneRanger \cite{droneranger} architecture by swapping the backbone for a ResNet and adding a bounding box feature gate to balance the weighting of image features and bounding box features for different distances.}
\label{fig:model}
\end{figure*}

Uncrewed aerial vehicles (UAVs), commonly referred to as drones, are now widely available and increasingly present in civilian airspace. This growth creates a practical need for robust perception methods that can detect drones at long range and estimate their distance to support tracking, safety monitoring, and situational awareness. Long-range drone perception is challenging because targets are often small, low-contrast, and affected by atmospheric conditions, motion blur, sensor noise, and background clutter (e.g., clouds, birds, and urban structures) \cite{coluccia_2024_drone_vs_bird, zheng_2021_detfly}. These issues are amplified in real-world deployments that must operate across diverse viewpoints, environments, and imaging modalities, including visible-spectrum and thermal imagery \cite{peterson2026_LRDDv3}.

While drone detection has received significant attention, accurate \emph{range estimation} from monocular imagery remains comparatively underexplored. Estimating distance is difficult because the target may occupy only a few pixels at long range, and scale cues can be ambiguous without calibrated geometry or multi-view observations \cite{li_2020_multiview_drone_tracking, uav-detect-pfiqs_dataset}. In practice, systems often rely on bounding box measurements and cropped appearance cues, both of which can be noisy: bounding boxes may be imperfect due to annotation or detector errors, and image crops may lose discriminative details as distance increases \cite{droneranger}. Consequently, models trained under narrow operating conditions may not generalize well across the wider range of distances and scene variability encountered in real-world data.

In this work, we focus on vision-based drone distance estimation using monocular images with bounding-box information. Building on prior work that combines cropped target appearance with bounding-box features, we evaluate architectural and training design choices that influence range prediction accuracy, including backbone selection, loss functions, and input resolution. We conduct experiments on a long-range drone dataset with accurate distance metadata and analyze performance trends across distance regimes, highlighting where range estimation is reliable and where it degrades due to target scale, data imbalance, and label or crop noise.

Our main contributions in this work are as follows: 
\begin{itemize}
    \item We propose DroneDAR (\textbf{Drone} \textbf{D}etection \textbf{A}nd \textbf{R}anging), a novel model for estimating the distance of drones that employs a custom bounding-box feature gate to improve robustness across distance regimes.
    \item We perform controlled experiments analyzing the impact of backbone capacity, loss function selection, and crop resolution on range estimation performance, and we report failure modes that arise at long distances.
\end{itemize}

\section{Related Work}\label{sec:related}

\subsection{Vision-Based Drone Detection and Long-Range Small-Object Perception}\

Vision-based UAV detection has been studied in a range of settings, including cooperative tracking applications and micro-UAV detection with distance estimation \cite{vision_uav_detection, vision_distance_muav}. Related work also examines detection performance limits and complementary sensing in drone-to-drone or sensor-fusion settings \cite{how_far_detection}. Other approaches rely heavily on state of the art object detection models, such as YOLO \cite{yolo} and Deformable-DETR \cite{zhu2020deformable}. Methods such as SAHI \cite{sahi} expand on that approach by enabling more effective use of image resolution by avoiding image downsizing during detection. Despite progress, long-range drone detection remains difficult because targets often occupy only a small number of pixels and may be visually similar to distractors such as birds. Recent benchmarking efforts emphasize these failure modes and highlight the need for careful evaluation under realistic conditions \cite{coluccia_2024_drone_vs_bird, zheng_2021_detfly}.

\subsection{Monocular Distance Estimation}

Estimating distance from monocular imagery is inherently ambiguous without explicit geometry or multi-view constraints. Recent approaches rely on large-scale training for generic monocular depth estimation and then fine-tune on metric depth information to provide final predictions \cite{depth_anything_v2, hu2024metric3dv2, depthpro}. However, these methods rely on limited metric depth data, and are often constrained to close-range applications.

Compared to detection, drone-specific distance estimation has received less attention. Recently, some works have explored UAV pose estimation at close range \cite{vision_pose_estimation, vision_based_formation}, and others have considered distance estimation for micro aerial vehicles using image evidence \cite{vision_distance_muav}. For small objects at long range, monocular range estimation is especially challenging because both appearance and scale cues degrade rapidly as distance increases, making predictions sensitive to crop quality, resolution, and label noise. Because of these challenges, distance estimation for drones at long-range remains a largely unexplored topic, and the single representative model for the task is DroneRanger, which estimates drone distance from monocular imagery using a combination of cropped target appearance and bounding-box information \cite{droneranger}.

\subsection{Gated Units for Multimodal Input}

Drone distance estimation is unique among other vision-based tasks, as instead of working with a purely image-based input, models also include bounding box features, resulting in a multi-modal input. For multi-modal inputs, it is important to balance the influence of the different modalities, and several works have explored methods to accomplish this. One of the main works is in Gated Multimodal Units (GMUs) \cite{arevalo2017gatedmultimodalunitsinformation}, which use multiplicative gates to learn how modalities should influence each other. Other feature fusion approaches use similar gate structures, but with alterations to account for different inputs or outputs \cite{featurefusion, gatedattentionfeature}. Based on those works, we develop our own feature gate unit specifically for balancing the impact of bounding box features and image crop features for drone range estimation.

\subsection{Drone Detection Datasets}

A practical limitation in long-range drone range estimation is the scarcity of publicly documented datasets that include accurate distance metadata alongside visual annotations. While many datasets include robust UAV detection or tracking information \cite{li_2020_multiview_drone_tracking, uav-detect-pfiqs_dataset}, very few datasets include any flight metadata or range information. The Multi-Sensor drone detection dataset includes rough categorical labels for drone distance, but lacks the precision required for distance estimation
\cite{svanstrom_2021_multisensor}. The DroneRanger paper solved this problem by creating a custom dataset in simulation, but they did not release it to the public. Since the release of DroneRanger, the Long-Range Drone Detection (LRDD) \cite{rouhi2024_LRDD} dataset has been released, and LRDDv2 \cite{rouhi2024_LRDDv2} was the first version of the dataset to include range information for approximately 8k instances. Most recently LRDDv3 \cite{peterson2026_LRDDv3} was released, containing over 100k images, with comprehensive range information across the entire dataset and official benchmark splits. As such, we use LRDDv3 for both training and evaluation in our work.

\section{Proposed Method}\label{sec:method}

\subsection{Model Architecture}

Our method builds on the DroneRanger architecture \cite{droneranger}, as it is one of the only current models for drone distance prediction. The DroneRanger model used custom CNN backbone, and also included bounding box features as input to a final regression head for predicting drone distance. 
DroneRanger also tested the impact of using both MSE loss and Huber loss, and found that Huber loss gave them an improvement of approximately 2 meters RMSE over using MSE loss.

Based on these findings, we propose an improved model that builds on the DroneRanger architecture, but makes several key changes to improve performance. First, we swap out the custom CNN backbone for a ResNet, and explore the impact of different backbone sizes and input resolutions. We test using both MSE and Huber loss to verify if the improvement shown by DroneRanger remains for changes to the model backbone. We also add a custom bounding box feature gate to more effectively control the impact of image features in the final prediction. An overview of our model architecture can be seen in Figure \ref{fig:model}.

We evaluate the impact of input resolution as it has a nuanced effect on drone distance estimation. Unlike typical vision tasks where higher resolution generally improves performance, drone crops exhibit extreme size variation, with close-range drones spanning hundreds of pixels while long-range targets occupy fewer than ten. Upscaling very small crops to large input dimensions introduces interpolation artifacts and redundant information that can mislead the network. Conversely, downscaling large crops may discard fine-grained features useful for nearby estimation. Our experiments (Section 4.3) empirically validate this tradeoff.

\begin{figure}[tbp]
\centering
\includegraphics[width=0.38\textwidth]{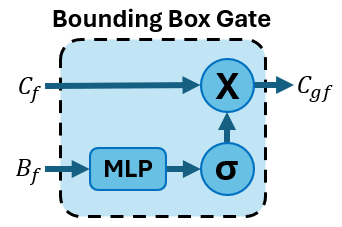}
\caption{The bounding box feature gate takes the bounding box features \(B_f\) and image crop features \(C_f\), and weights \(C_f\) based on \(B_f\) to balance the tradeoff of image and bounding box features for different distances.}
\label{fig:bbox_gate}
\end{figure}

In addition to changing the model backbone and exploring different input resolutions, we develop a novel bounding box feature gate to balance the impact of bounding box features and image crop features. This allows the model to learn when image crop features are reliable, and when it is more useful to depend on bounding box features. This is especially important for long-range drone distance estimation, as at longer ranges there are very few pixels representing the drone, but at closer ranges the image features include more information to improve prediction accuracy. As shown in figure \ref{fig:bbox_gate}, the bounding box gate takes the bounding box features \(B_f\) (e.g. area, aspect ratio) as an input to a dense MLP layer that projects them to the same dimension as the cropped image feature output of the ResNet backbone \(C_f\). The result is then fed into a sigmoid activation function, and multiplied with the image features \(C_f\) to produce the gated feature output \(C_gf\). 
The feature gate allows the model to weight the features of the image crop based on the bounding box features by learning the weights for the MLP projection layer. This allows the model to balance its focus between features depending on the quality of the image input

\subsection{Dataset and Training}

As the DroneRanger paper did not make their dataset publicly available, we instead use the LRDDv3 \cite{peterson2026_LRDDv3} dataset. LRDDv3 consists of 102k RGB images, captured at 4k resolution and sub-sampled at 5FPS from flight video, along with drone distance information. For training, we use the official splits of LRDDv3, and use ground-truth bounding box labels for drone image crops and bounding box features for our model. All models were trained on an AMD Ryzen Threadripper 3960X 24-Core Processor and an NVIDIA GeForce RTX 3090 GPU for 50 epochs using the Adam optimizer with a learning rate of 1e-3. Different model backbones used different batch sizes to accommodate for available GPU memory: DroneRanger and ResNet18 backbones used 64 batch size, while ResNet50 backbones used a batch size of 32. For testing, the best weights from training were used.

\section{Experimental Results}\label{sec:experiments}

\subsection{Backbone Type}

\begin{table*}[tbp]
\caption{Results of range prediction models using different backbones. Models were trained using Huber loss and 512 resolution crop, with the ResNet models outperforming the DroneRanger baseline. \label{table:backbone_results}}
\centering
\begin{tabular}{l|cc|cc|cc|cc}
\textbf{BACKBONE} &
  \multicolumn{2}{c|}{\textbf{Full range}} &
  \multicolumn{2}{c|}{\textbf{0-300 ft}} &
  \multicolumn{2}{c|}{\textbf{0-200 ft}} &
  \multicolumn{2}{c}{\textbf{0-100 ft}} \\ \hline
\textbf{Model}      & \textbf{MAE}   & \textbf{RMSE}  & \textbf{MAE}   & \textbf{RMSE}  & \textbf{MAE}   & \textbf{RMSE}  & \textbf{MAE}   & \textbf{RMSE}  \\ \hline
DroneRanger & 35.95 & 70.67 & 27.52 & 42.65 & 24.31 & 35.13 & 18.15 & 27.40 \\ 
ResNet18  & \textit{33.96} & \textit{70.12} & \textit{25.41} & \textit{41.58} & \textit{21.90} & \textit{33.02} & \textit{15.49} & \textit{23.89} \\ 
ResNet50    & \textbf{33.39} & \textbf{69.99}          & \textbf{24.71} & \textbf{40.68} & \textbf{21.09} & \textbf{31.61} & \textbf{14.93} & \textbf{23.24} \\ 
\end{tabular}
\end{table*}

Table \ref{table:backbone_results} shows the results of swapping the original DroneRanger backbone with different ResNet models when tested on the LRDDv3 test set. Both ResNet models improved on the DroneRanger backbone by $\sim$2\% MAE and $\sim$0.5\% RMSE overall, with substantially larger improvements a closer ranges. While both ResNet backbones outperformed the DroneRanger baseline, the differences between ResNet backbones were very small, indicating that changes to backbone size offer minimal advantages after a baseline size has been reached.
It is important to note that the DroneRanger model code was not made publicly available, so we recreated the model architecture based on the information provided in the paper for our comparisons.

\subsection{Huber vs. MSE Loss}

\begin{table*}[tbp]
\caption{Results for models trained using MSE loss and Huber loss.  For longer ranges MSE produced a better RMSE result, but Huber loss produced better results for RMSE at close range and MAE across all distances. \label{table:loss_results}}
\centering
\begin{tabular}{l|cc|cc|cc|cc}
\textbf{LOSS} &
  \multicolumn{2}{c|}{\textbf{Full range}} &
  \multicolumn{2}{c|}{\textbf{0-300 ft}} &
  \multicolumn{2}{c|}{\textbf{0-200 ft}} &
  \multicolumn{2}{c}{\textbf{0-100 ft}} \\ \hline
\textbf{Model}      & \textbf{MAE}   & \textbf{RMSE}  & \textbf{MAE}   & \textbf{RMSE}  & \textbf{MAE}   & \textbf{RMSE}  & \textbf{MAE}   & \textbf{RMSE}  \\ \hline
DroneRanger (MSE)   & 36.34          & \textbf{69.92} & 28.06          & \textbf{42.62} & 24.93          & 35.28          & 19.96          & 29.70          \\
DroneRanger (Huber) & \textbf{35.95} & 70.67          & \textbf{27.52} & 42.65          & \textbf{24.31} & \textbf{35.13} & \textbf{18.15} & \textbf{27.40} \\ \hline
ResNet18 (MSE)      & 34.71          & \textbf{69.07} & 26.35          & \textbf{41.16} & 23.06          & 33.33          & 17.86          & 26.25          \\
ResNet18 (Huber)    & \textbf{33.96} & 70.12          & \textbf{25.41} & 41.58          & \textbf{21.90} & \textbf{33.02} & \textbf{15.49} & \textbf{23.89} \\ \hline
ResNet50 (MSE)      & 35.01          & \textbf{68.06} & 26.93          & 41.68          & 23.56          & 33.69          & 19.09          & 27.96          \\
ResNet50 (Huber)    & \textbf{33.39} & 69.99          & \textbf{24.71} & \textbf{40.68} & \textbf{21.09} & \textbf{31.61} & \textbf{14.93} & \textbf{23.24} \\ 
\end{tabular}
\end{table*}

Table \ref{table:loss_results} shows the results of training different models using both MSE loss and Huber loss. For longer ranges MSE produced a better RMSE result, but Huber loss produced better results for RMSE at close range and MAE across all distances. This was slightly different from the findings of the DroneRanger paper, which reported large improvements in RMSE by using Huber loss; however the end result was still that Huber loss produced better results overall. As such we use Huber loss for training subsequent models.

\subsection{Input Type and Resolution}

\begin{table*}[tbp]
\caption{Results for different image crop input resolutions. In general, lower input resolutions yield better results as they do not overly-upscale small drone crops. However, for the ResNet50 model at close range, the larger input resolution results in the best performance. \label{table:resolution_results}}
\centering
\begin{tabular}{l|cc|cc|cc|cc}
\textbf{RESOLUTION} &
  \multicolumn{2}{c|}{\textbf{Full range}} &
  \multicolumn{2}{c|}{\textbf{0-300 ft}} &
  \multicolumn{2}{c|}{\textbf{0-200 ft}} &
  \multicolumn{2}{c}{\textbf{0-100 ft}} \\ \hline
\textbf{Model} & \textbf{MAE}   & \textbf{RMSE}  & \textbf{MAE}   & \textbf{RMSE}  & \textbf{MAE}   & \textbf{RMSE}  & \textbf{MAE}   & \textbf{RMSE}  \\ \hline
ResNet18 220 &
  \textbf{32.84} &
  \textbf{68.68} &
  \textbf{24.38} &
  \textbf{40.18} &
  \textbf{21.10} &
  \textbf{32.42} &
  \textbf{14.96} &
  \textbf{23.80} \\
ResNet18 360   & 34.19          & \textit{69.44} & 25.86          & 42.28          & 22.43          & 34.28          & 15.96          & 24.66          \\
ResNet18 512   & \textit{33.96} & 70.12          & \textit{25.41} & \textit{41.58} & \textit{21.90} & \textit{33.02} & \textit{15.49} & \textit{23.89} \\ \hline
ResNet50 220   & \textbf{32.96} & \textbf{69.46} & \textbf{24.46} & \textit{41.18} & \textbf{21.05} & 33.11          & \textit{15.34} & 25.05          \\
ResNet50 360   & 33.61          & \textit{69.69} & 25.11          & 41.24          & 21.72          & \textit{33.09} & 15.75          & \textit{24.66} \\
ResNet50 512   & \textit{33.39} & 69.99          & \textit{24.71} & \textbf{40.68} & \textit{21.09} & \textbf{31.61} & \textbf{14.93} & \textbf{23.24}
\end{tabular}
\end{table*}

Table \ref{table:resolution_results} shows the results of different image input resolutions on the final model results. While for most generic computer vision applications an increase in input resolution provides better results, drone range estimation is unique in that the actual resolution of drone crops can vary widely based on drone distance. Because of this, our resizing of drone image crops to a fixed input size can have different effects depending on the range of the drone. For long-range drones, having a large input size upscales very few pixels into a very large image, with much redundant information, and for close-range drones, having a small input size downscales image crops, potentially losing information. 
In table \ref{table:resolution_results} we test using 220x220, 360x360, and 512x512 pixel input resolutions for two different ResNet backbones. We see that for longer ranges, using a smaller input resolution increases performance, as small image crops are not overly upscaled. For the ResNet50 backbone, the 512 resolution has better performance at closer ranges, as image features are not lost due to downscaling. However, for the ResNet18 backbone, the 220 input resolution remains the best performer for close ranges as well, indicating that the extra resolution may not be needed for accurate close-range distance estimation.

\begin{table*}[tbp]
\caption{Results for adding the bounding box feature gate. The addition of the gate offers small improvements overall, with most of the benefit at longer ranges where drone image features offer less information. \label{table:gate_results}}
\centering
\begin{tabular}{l|cc|cc|cc|cc}
\textbf{BBOX GATE} &
  \multicolumn{2}{c|}{\textbf{Full range}} &
  \multicolumn{2}{c|}{\textbf{0-300 ft}} &
  \multicolumn{2}{c|}{\textbf{0-200 ft}} &
  \multicolumn{2}{c}{\textbf{0-100 ft}} \\ \hline
\textbf{Model}     & \textbf{MAE}   & \textbf{RMSE}  & \textbf{MAE}   & \textbf{RMSE}  & \textbf{MAE}   & \textbf{RMSE}  & \textbf{MAE}   & \textbf{RMSE}  \\ \hline
ResNet18 220       & 32.84          & 68.68          & 24.38          & 40.18          & 21.10          & 32.42          & \textbf{14.96} & \textbf{23.80} \\
ResNet18 220 Gated & \textbf{32.39} & \textbf{68.49} & \textbf{23.89} & \textbf{39.52} & \textbf{20.51} & \textbf{31.07} & 15.23          & 24.30          \\ \hline
ResNet18 512       & 33.96          & 70.12          & 25.41          & 41.58          & 21.90          & 33.02          & \textbf{15.49} & \textbf{23.89} \\
\multicolumn{1}{c|}{ResNet18 512 Gated} &
  \textbf{33.66} &
  \textbf{69.72} &
  \textbf{25.16} &
  \textbf{41.23} &
  \textbf{21.66} &
  \textbf{32.52} &
  15.52 &
  24.09 \\
\end{tabular}
\end{table*}

\begin{table*}[tbp]
\caption{Comparison of our final DroneDAR model with DroneRanger. Our model outperforms DroneRanger in both MAE and RMSE across all distances. 
\label{table:final_comparison}}
\centering
\begin{tabular}{l|cc|cc|cc|cc}
\textbf{FINAL MODEL} &
  \multicolumn{2}{c|}{\textbf{Full range}} &
  \multicolumn{2}{c|}{\textbf{0-300 ft}} &
  \multicolumn{2}{c|}{\textbf{0-200 ft}} &
  \multicolumn{2}{c}{\textbf{0-100 ft}} \\ \hline
\textbf{Model} & \textbf{MAE} & \textbf{RMSE} & \textbf{MAE} & \textbf{RMSE} & \textbf{MAE} & \textbf{RMSE} & \textbf{MAE} & \textbf{RMSE} \\ \hline
DroneRanger    & 35.95        & 70.67         & 27.52        & 42.65         & 24.31        & 35.13         & 18.15        & 27.40         \\
DroneDAR (Ours) &
  \textbf{32.39} &
  \textbf{68.49} &
  \textbf{23.89} &
  \textbf{39.52} &
  \textbf{20.51} &
  \textbf{31.07} &
  \textbf{15.23} &
  \textbf{24.30}
\end{tabular}
\end{table*}

\subsection{Bounding Box Feature Gate}

Table \ref{table:gate_results} shows the results of adding the bounding box feature gate to the model. For both of the ResNet18 models the addition of the feature gate improved the overall performance of the model, especially at longer ranges. However, for very close ranges the performance decreased slightly. 
This could be due to a variety of factors, such as drone crops reaching a high enough resolution that they are downscaled instead of upscaled to reach the fixed-size model input, or that by improving long-range detection the model failed to generalize as well to close range estimation. As such, this is a potential avenue for future work.


\subsection{Final Result Comparison}



Table \ref{table:final_comparison} shows the comparison between our final DroneDAR model  with the DroneRanger model. Our model outperforms the DroneRanger model by 
3.56 feet MAE and 2.18 feet RMSE across the full LRDDv3 test set, with consistently better performance across all distances.

\begin{figure*}[tbp] 
\centering
\includegraphics[width=0.71\textwidth]{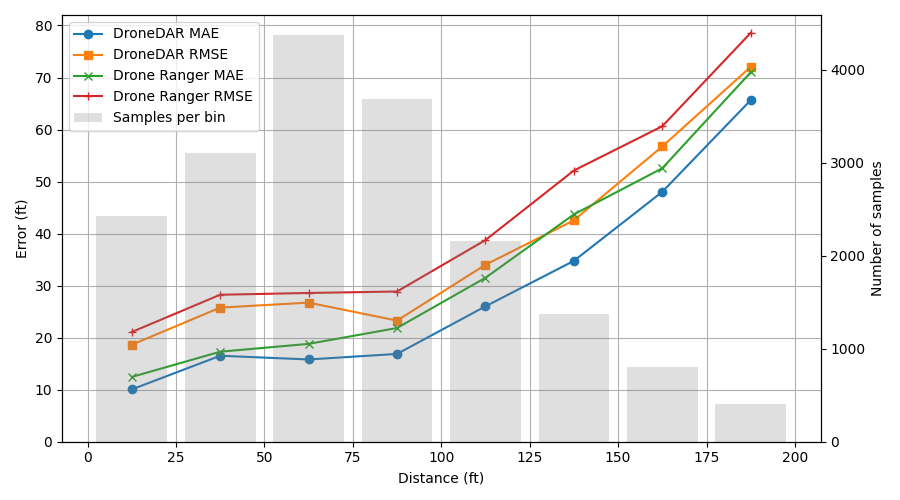}
\caption{Plot of the error metrics for our final model compared to the original DroneRanger model for distances up to 200 ft. Longer distances are not shown as data is concentrated under 200 ft, and the closer view highlights model differences. Our model consistently outperforms the DroneRanger model in both MAE and RMSE for all distances.}
\label{fig:droneranger_resnet18_gated_error_plot}
\end{figure*}

Figure \ref{fig:droneranger_resnet18_gated_error_plot} shows the plot of the error metrics for our final model compared to the DroneRanger model for distances up to 200 feet. We show a zoomed-in view for figure clarity, and the trends shown in the figure hold for longer ranges.
As shown in the figure, our model consistently outperforms DroneRanger in both MAE and RMSE for all distances (including above 200 feet). Both models have the best performance at closer range, especially below 100 feet where the most data is available. Above a distance of 100 feet the error begins to increase, likely due to a combination of decreasing drone crop size and available image features, as well as the amount of training data.

\begin{figure*}[tbp] 
\centering
\includegraphics[width=0.95\textwidth]{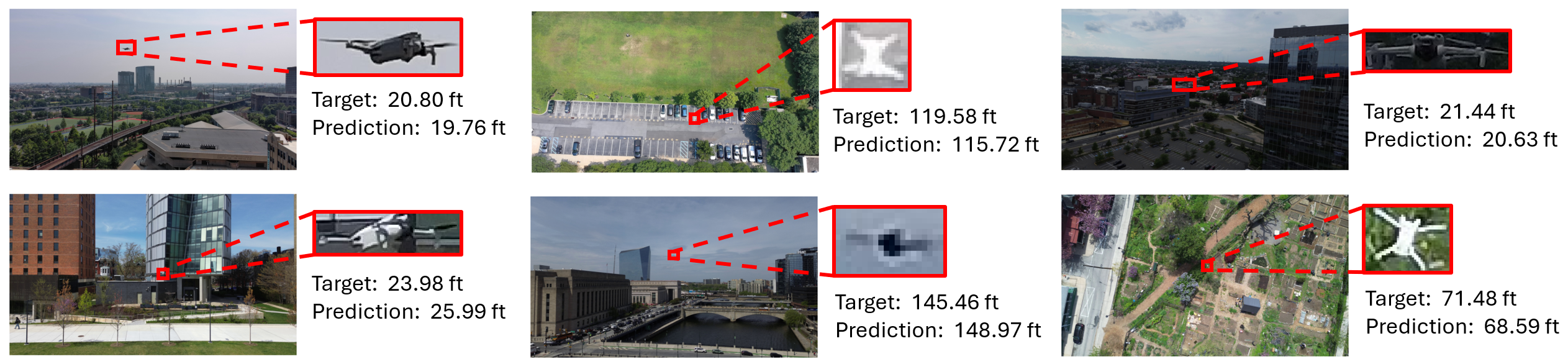}
\caption{Examples of success cases where the model very accurately predicted the drone distance. These covered a wide variety of camera angles, lighting conditions, and backgrounds.}
\label{fig:good_predictions}
\end{figure*}

Figure \ref{fig:good_predictions} shows examples of cases where our final model predicted drone distances very accurately. These cases include a wide variety of camera angles, lighting conditions, and backgrounds. Generally these cases also had accurate ground truth bounding boxes, which contributed to more accurate results.

\begin{figure*}[t] 
\centering
\includegraphics[width=0.95\textwidth]{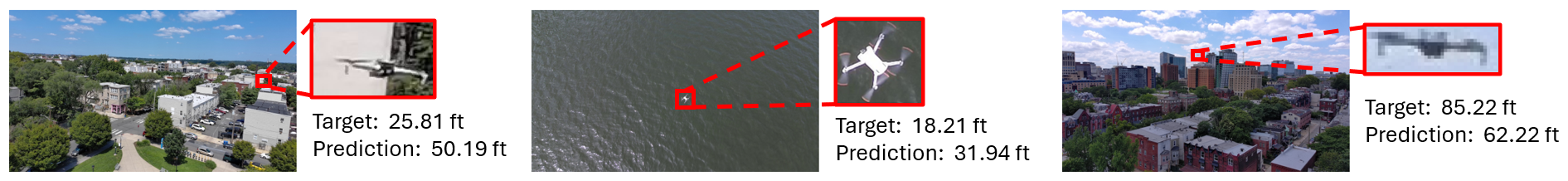}
\caption{Examples of failure cases where the model predicted inaccurate distances. Some of the failures were the result of inaccurate or noisy bounding boxes (Left: inaccurate box; Middle: larger bounding box size from propellers) 
Other cases had accurate bounding boxes, but still had inaccurate predictions (Right).}
\label{fig:bad_predictions}
\end{figure*}

Figure \ref{fig:bad_predictions} shows examples of failure cases where model produced inaccurate distance predictions. Some failures were likely impacted by noisy bounding boxes such as in cases shown in the top left and bottom left of the figure. Other cases had predictions with larger error values even with very accurate bounding boxes, as shown in the right side of the figure. This is not surprising, due to the inherent difficulty of the distance prediction task, especially when drones are viewed at such a wide variety of angles and in different lighting conditions and backgrounds.

\subsection{Discussion}

Across the ablations in previous sections, several trends emerged. First, increasing backbone size provided only marginal improvements: ResNet-18 and ResNet-50 improved substantially on the DroneRanger baseline, but performed similarly to each other, indicating that model capacity saturates quickly for this task. This likely occurs because the information content in many crops is inherently limited, so larger networks do not have additional signal to exploit.

Second, input resolution had a non-trivial tradeoff. Higher resolutions can preserve detail for close-range drones, but for long-range drones they primarily upscale a very small number of pixels into a large input, introducing redundancy and interpolation noise. Empirically, a smaller crop size performed best overall, and the best-performing configuration in our study was a ResNet-18 backbone with a $220 \times 220$ crop resolution trained with Huber loss.

Third, adding the proposed bounding-box feature gate produced additional gains in overall performance, particularly at longer ranges. This suggests that explicitly learning how much to rely on geometric cues (bounding-box features) versus appearance cues (image features) is beneficial when the visual crop becomes unreliable. However, the failure examples also highlight an important limitation: when the bounding box is inaccurate or noisy, both the crop and the derived box features are corrupted, and the model can produce large range errors. Improving robustness to imperfect detections is therefore a promising direction for future work. 

\section{Conclusion}\label{sec:conclusion}

In this work we introduced DroneDAR, a method for monocular drone distance estimation which builds on a DroneRanger baseline by replacing the custom CNN backbone with a ResNet and incorporating a lightweight bounding-box feature gate to better balance appearance and geometric cues. Experiments on LRDDv3 show that DroneDAR improves over the DroneRanger baseline across all distance regimes. Our ablations further indicate that (i) backbone capacity saturates quickly for long-range small-object crops, (ii) crop resolution must be chosen carefully due to the severe upscaling/downscaling tradeoff, and (iii) explicitly gating image features using bounding-box cues can provide consistent gains, especially at longer ranges. Despite these improvements, performance remains limited at extreme distances where targets occupy only a few pixels and bounding-box noise becomes a dominant error source. Future work should focus on improving robustness to imperfect detections and leveraging additional cues such as temporal information to stabilize long-range predictions.








\bibliographystyle{IEEEtran}
\bibliography{ref.bib}

@INPROCEEDINGS{droneranger,
  author={Azad, Hamid and Mehta, Varun and Mantegh, Iraj and Bolic, Miodrag},
  booktitle={2024 International Conference on Unmanned Aircraft Systems (ICUAS)}, 
  title={DroneRanger: Vision-Driven Deep Learning for Drone Distance Estimation}, 
  year={2024},
  volume={},
  number={},
  doi={10.1109/ICUAS60882.2024.10557013}
}

@Article{vision_uav_detection,
AUTHOR = {Opromolla, Roberto and Fasano, Giancarmine and Accardo, Domenico},
TITLE = {A Vision-Based Approach to UAV Detection and Tracking in Cooperative Applications},
JOURNAL = {Sensors},
VOLUME = {18},
YEAR = {2018},
NUMBER = {10},
ARTICLE-NUMBER = {3391},
URL = {https://www.mdpi.com/1424-8220/18/10/3391},
PubMedID = {30309035},
ISSN = {1424-8220},
DOI = {10.3390/s18103391}
}

@Article{vision_distance_muav,
AUTHOR = {Gökçe, Fatih and Üçoluk, Göktürk and Şahin, Erol and Kalkan, Sinan},
TITLE = {Vision-Based Detection and Distance Estimation of Micro Unmanned Aerial Vehicles},
JOURNAL = {Sensors},
VOLUME = {15},
YEAR = {2015},
NUMBER = {9},
PAGES = {23805--23846},
URL = {https://www.mdpi.com/1424-8220/15/9/23805},
PubMedID = {26393599},
ISSN = {1424-8220},
DOI = {10.3390/s150923805}
}

@inproceedings{how_far_detection,
author = {Kim, Juann and Kim, Youngseo and Shin, Heeyeon and Wang, Mia and Matson, Eric},
booktitle={15th International Conference on Agents and Artificial Intelligence (ICAART 2023)},
year = {2023},
month = {01},
pages = {877-884},
title = {How Far Can a Drone be Detected? A Drone-to-Drone Detection System Using Sensor Fusion},
doi = {10.5220/0011791000003393}
}

@INPROCEEDINGS{vision_based_formation,
  author={Lin, Feng and Peng, Kemao and Dong, Xiangxu and Zhao, Shiyu and Chen, Ben M.},
  booktitle={11th IEEE International Conference on Control \& Automation (ICCA)}, 
  title={Vision-based formation for UAVs}, 
  year={2014},
  volume={},
  number={},
  pages={1375-1380},
  doi={10.1109/ICCA.2014.6871124}}

@INPROCEEDINGS{vision_pose_estimation,
  author={Zhang, Mengmi and Lin, Feng and Chen, Ben M.},
  booktitle={2014 13th International Conference on Control Automation Robotics \& Vision (ICARCV)}, 
  title={Vision-based detection and pose estimation for formation of micro aerial vehicles}, 
  year={2014},
  volume={},
  number={},
  pages={1473-1478},
  doi={10.1109/ICARCV.2014.7064533}}

@article{svanstrom_2021_multisensor,
title = {A dataset for multi-sensor drone detection},
journal = {Data in Brief},
volume = {39},
pages = {107521},
year = {2021},
issn = {2352-3409},
doi = {https://doi.org/10.1016/j.dib.2021.107521},
url = {https://www.sciencedirect.com/science/article/pii/S2352340921007976},
author = {Fredrik Svanström and Fernando Alonso-Fernandez and Cristofer Englund},
}

@ARTICLE{zheng_2021_detfly,
  author={Zheng, Ye and Chen, Zhang and Lv, Dailin and Li, Zhixing and Lan, Zhenzhong and Zhao, Shiyu},
  journal={IEEE Robotics and Automation Letters}, 
  title={Air-to-Air Visual Detection of Micro-UAVs: An Experimental Evaluation of Deep Learning}, 
  year={2021},
  volume={6},
  number={2},
  pages={1020-1027},
  doi={10.1109/LRA.2021.3056059}}

@ARTICLE{coluccia_2024_drone_vs_bird,
  author={Coluccia, Angelo and Fascista, Alessio and Sommer, Lars and Schumann, Arne and Dimou, Anastasios and Zarpalas, Dimitrios},
  journal={IEEE Open Journal of Signal Processing}, 
  title={The Drone-vs-Bird Detection Grand Challenge at ICASSP 2023: A Review of Methods and Results}, 
  year={2024},
  volume={5},
  number={},
  pages={766-779},
  doi={10.1109/OJSP.2024.3379073}}

@INPROCEEDINGS{rouhi2024_LRDD,
  author={Rouhi, Amirreza and Umare, Himanshu and Patal, Sneh and Kapoor, Ritik and Deshpande, Namit and Arezoomandan, Solmaz and Shah, Princie and Han, David},
  booktitle={2024 IEEE International Conference on Consumer Electronics (ICCE)}, 
  title={Long-Range Drone Detection Dataset}, 
  year={2024},
  volume={},
  number={},
  doi={10.1109/ICCE59016.2024.10444135}
}

@INPROCEEDINGS{rouhi2024_LRDDv2,
  author={Rouhi, Amirreza and Patel, Sneh and McCarthy, Noah and Khan, Siddiqa and Khorsand, Hadi and Lefkowitz, Kaleb and Han, David},
  booktitle={2024 International Symposium of Robotics Research (ISRR)}, 
  title={LRDDv2: Enhanced Long-Range Drone Detection Dataset with Range Information and Comprehensive Real-World Challenges}, 
  year={2024},
  volume={},
  number={}
}

@article{yolo,
  author       = {Joseph Redmon and
                  Santosh Kumar Divvala and
                  Ross B. Girshick and
                  Ali Farhadi},
  title        = {You Only Look Once: Unified, Real-Time Object Detection},
  journal      = {CoRR},
  volume       = {abs/1506.02640},
  year         = {2015},
  url          = {http://arxiv.org/abs/1506.02640},
  eprinttype    = {arXiv},
  eprint       = {1506.02640},
  bibsource    = {dblp computer science bibliography, https://dblp.org}
}

@article{zhu2020deformable,
  title={Deformable DETR: Deformable Transformers for End-to-End Object Detection},
  author={Zhu, Xizhou and Su, Weijie and Lu, Lewei and Li, Bin and Wang, Xiaogang and Dai, Jifeng},
  journal={arXiv preprint arXiv:2010.04159},
  year={2020}
}

@inproceedings{sahi,
   title={Slicing Aided Hyper Inference and Fine-Tuning for Small Object Detection},
   url={http://dx.doi.org/10.1109/ICIP46576.2022.9897990},
   DOI={10.1109/icip46576.2022.9897990},
   booktitle={2022 IEEE International Conference on Image Processing (ICIP)},
   publisher={IEEE},
   author={Akyon, Fatih Cagatay and Onur Altinuc, Sinan and Temizel, Alptekin},
   year={2022},
   }

@article{depth_anything_v2,
  title={Depth Anything V2},
  author={Yang, Lihe and Kang, Bingyi and Huang, Zilong and Zhao, Zhen and Xu, Xiaogang and Feng, Jiashi and Zhao, Hengshuang},
  journal={arXiv:2406.09414},
  year={2024}
}

@article{hu2024metric3dv2,
  title={Metric3d v2: A versatile monocular geometric foundation model for zero-shot metric depth and surface normal estimation},
  author={Hu, Mu and Yin, Wei and Zhang, Chi and Cai, Zhipeng and Long, Xiaoxiao and Chen, Hao and Wang, Kaixuan and Yu, Gang and Shen, Chunhua and Shen, Shaojie},
  journal={IEEE Transactions on Pattern Analysis and Machine Intelligence},
  year={2024},
  publisher={IEEE}
}

@inproceedings{depthpro,
  author     = {Aleksei Bochkovskii and Ama\"{e}l Delaunoy and Hugo Germain and Marcel Santos and
               Yichao Zhou and Stephan R. Richter and Vladlen Koltun},
  title      = {Depth Pro: Sharp Monocular Metric Depth in Less Than a Second},
  booktitle  = {International Conference on Learning Representations},
  year       = {2025},
  url        = {https://arxiv.org/abs/2410.02073},
}

@INPROCEEDINGS{li_2020_multiview_drone_tracking,
  author={Li, Jingtong and Murray, Jesse and Ismaili, Dorina and Schindler, Konrad and Albl, Cenek},
  booktitle={2020 IEEE/RSJ International Conference on Intelligent Robots and Systems (IROS)}, 
  title={Reconstruction of 3D flight trajectories from ad-hoc camera networks}, 
  year={2020},
  volume={},
  number={},
  pages={1621-1628},
  doi={10.1109/IROS45743.2020.9341479}}

@misc{
uav-detect-pfiqs_dataset,
title = { uav detect Dataset },
type = { Open Source Dataset },
author = { GET },
howpublished = { \url{ https://universe.roboflow.com/get/uav-detect-pfiqs } },
url = { https://universe.roboflow.com/get/uav-detect-pfiqs },
journal = { Roboflow Universe },
publisher = { Roboflow },
year = { 2023 },
month = { jan },
note = { visited on 2025-09-15 },
}

@misc{arevalo2017gatedmultimodalunitsinformation,
      title={Gated Multimodal Units for Information Fusion}, 
      author={John Arevalo and Thamar Solorio and Manuel Montes-y-Gómez and Fabio A. González},
      year={2017},
      eprint={1702.01992},
      archivePrefix={arXiv},
      primaryClass={stat.ML},
      url={https://arxiv.org/abs/1702.01992}, 
}

@INPROCEEDINGS{featurefusion,
  author={Sun, Zepeng and Jin, Dongyin and Deng, Jian and Zhang, Mengyang and Shao, Zhenzhou},
  booktitle={IEEE International Conference on Robotics and Biomimetics}, 
  title={Feature Fusion Module Based on Gate Mechanism for Object Detection}, 
  year={2023},
  volume={},
  number={},
  doi={10.1109/ROBIO58561.2023.10354575}}

@InProceedings{gatedattentionfeature,
author="Ramzan, Muhammad Umer
and Khaddim, Wahab
and Rana, Muhammad Ehsan
and Ali, Usman
and Ali, Manohar
and ul Hassan, Fiaz
and Mehmood, Fatima",
editor="Bhateja, Vikrant
and Abdul Hameed, Vazeerudeen
and Udgata, Siba K.
and Azar, Ahmad Taher",
title="Gated-Attention Feature-Fusion Based Framework for Poverty Prediction",
booktitle="Innovations in Communication Networks: Sustainability for Societal and Industrial Impact",
year="2025",
publisher="Springer Nature Singapore",
address="Singapore",
pages="415--426",
}

@INPROCEEDINGS{peterson2026_LRDDv3,
  author={Peterson, Knut and Mayers, Zaid and Yousuf, Azmain and Chowdhury, Priontu and Zaczepinski, Asher and Arezoomandan, Solmaz and Maarefdoust, Reihaneh and Han, David},
  booktitle={2026 IEEE International Conference on Robotics and Automation (ICRA)}, 
  title={LRDDv3: High-Resolution Long-Range Drone Detection Dataset with Range Information and Thermal Data}, 
  year={2026},
  volume={},
  number={}
}
\end{document}